\documentclass[journal]{IEEEtran}

\ifCLASSINFOpdf
\else
   \usepackage[dvips]{graphicx}
\fi
\usepackage{url}

\usepackage{graphicx}
\usepackage{cite}
\usepackage{amsmath,amssymb,amsfonts}
\usepackage{algorithmic}
\usepackage{textcomp}
\usepackage{xcolor}
\usepackage{array}
\usepackage{booktabs}
\usepackage{multirow}
\usepackage{url}
\usepackage{float}
\usepackage{newunicodechar}
\newunicodechar{‑}{\mbox{-}}

\begin{document}

\title{Spatial-Temporal Multi-scale Network for Screen Content Video Quality Enhancement}

\author{Ziyin Huang, Sik-Ho Tsang, Xinyuan Qin, Yui-Lam Chan, \IEEEmembership{Member, IEEE}, Xueling Zhou, and Feiyu Chen
\thanks{This work has been submitted to the IEEE for possible publication. Copyright may be transferred without notice, after which this version may no longer be accessible.}
\thanks{This work is supported by the Centre for Advances in Reliability and Safety (CAiRs) under Utilization of the CAiRs Secondment Fund (68BS), and Research Platform for Advanced Audio and Speech Signal Processing (P0049192) funded by Innovation Technology Co. Ltd. \textit{(Ziyin Huang and Sik-Ho Tsang contributed equally to this work.)(Corresponding author: Yui-Lam Chan.)}}
\thanks{Ziyin Huang is with School of Artificial Intelligence, Shenzhen Polytechnic University, Shenzhen, China, and with the Department of Electrical and Electronic Engineering, The Hong Kong Polytechnic University, Hong Kong, China (e-mail: ziyin.huang@connect.polyu.hk).}
\thanks{Sik-Ho Tsang is with the Department of Computer Science, Hong Kong Chu Hai College, Hong Kong, China (e-mail: harristsang@chuhai.edu.hk).}
\thanks{Xinyuan Qin and Yui-Lam Chan are with the Department of Electrical and Electronic Engineering, The Hong Kong Polytechnic University, Hong Kong, China (e-mail: littlepo.qin@connect.polyu.hk,  enylchan@polyu.edu.hk).}
\thanks{Xueling Zhou is with the College of Eilte Engineers, Dongguan University of Technology, Dongguan, China (e-mail: zhouxueling@dgut.edu.cn).}
\thanks{Feiyu Chen is with Department of Computer Science, City University of Hong Kong, Hong Kong, China (e-mail: feiyuchen3-c@my.cityu.edu.hk).}}
\markboth{}{}
\maketitle

\begin{abstract}
Different from natural videos, Screen Content Videos (SCVs) are characterized by abrupt motion, scene switches, and high-frequency details such as text and graphics. Conventional video enhancement methods, which rely heavily on temporal continuity, often suffer from performance degradation when processing SCVs due to the disruption of temporal correlations. To address these challenges, we propose the Spatial-Temporal Multi-scale Network (STM-Net), a novel framework specifically tailored for compressed SCV enhancement. Our approach integrates three complementary components: a Prior-Guided Spatio-Temporal Dispatcher (PG-STD) that routes input into three parallel streams to avoid feature contamination, a Bidirectional Temporal Feature Extraction (BTFE) module that adaptively handles abrupt transitions without explicit detection, and a Cascaded Multi-scale Feature Distillation (CMFD) module that preserves critical high-frequency details. Experimental results demonstrate that STM-Net outperforms state-of-the-art methods in both objective metrics and subjective visual quality, providing a robust solution for screen content artifacts. Code is available at https://github.com/HUANGZiyin1/STM-Net.
\end{abstract}


\IEEEpeerreviewmaketitle

\section{Introduction}

\IEEEPARstart{T}{he} rapid growth of cloud computing and remote collaboration has made Screen Content Video (SCV) transmission ubiquitous. Pre-COVID MOOCs and remote work drove demand for efficient coding of slides, code editors, and GUIs, amplified by pandemic videoconferencing and sustained in hybrid settings \cite{muller2023videoconf, ishii2023remoteworking}.

Various quality enhancement techniques have been proposed to improve visual fidelity for natural content \cite{wan2020bilateral, qiu2021lowlight, shi2022lowlight, ji2024sisr, cheng2026facesr}. However, unlike natural content, SCVs feature abrupt scene switches and computer-generated content, such as sharp edges, homogeneous regions, and repetitive patterns, which demands specialized compression like HEVC's Screen Content Coding (SCC) \cite{yu2014requirements, ma2014advanced, xu2015overview, xu2016intra, xu2022scc_overview}. Despite efficiency gains, compression artifacts such as blurring and ringing around text edges persist, severely impairing readability. Conventional enhancement methods fail due to abrupt motions (scrolling, scene switches) that disrupt temporal correlations where deep flow-based \cite{yang2018multi} or deformable convolution \cite{deng2020spatio} approaches propagate errors from unrelated frames, leading to severe visual artifacts \cite{huang2022qecf}.

Single‑frame methods \cite{park2016cnn, dai2016convolutional, dai2017convolutional, wang2017novel, yang2018enhancing}, such as IFCNN, VRCNN, DCAD, and QE-CNN, focus primarily on spatial information to restore compressed frames. These methods effectively exploit local cues but neglect temporal continuity, limiting their performance in video sequences.

Multi‑frame approaches \cite{yang2018multi, guan2019mfqe, deng2020spatio, luo2022coarse, luo2023spatio, zhu2024compressed, huang2022qecf, huang2024spatio, huang2024frame}, such as MFQE \cite{guan2019mfqe}, rely on optical flow alignment, which is prone to errors in SCVs due to large, discontinuous displacements. Deformable convolution networks like STDF \cite{deng2020spatio} and STDR \cite{luo2023spatio} offer flexible alignment but struggle with scene switches where reference frames differ entirely. Recent SCV‑specific methods, including TGAF \cite{zhu2024compressed} and QECF  \cite{huang2022qecf}, introduce cross‑frame fusion. However, these methods typically process temporal data in a single stream, which risk feature contamination during scene switches.

To address these issues, we propose the Spatial-Temporal Multi-scale Network (STM-Net) to address temporal discontinuity and spatial high-frequency loss with three innovations:
\begin{itemize}
    \item \textbf{Prior‑Guided Spatio‑Temporal Dispatcher (PG‑STD):} PG‑STD acts as a feature dispatcher that route input frames to three parallel streams (preceding, succeeding, current frames) to prevent scene-cut contamination.
    \item \textbf{Bidirectional Temporal Feature Extraction (BTFE):} Unlike single-stream methods, BTFE employs a dual-stream architecture with cross-connections to adaptively prioritize reliable temporal contexts, avoiding explicit scene detection.
    \item \textbf{Cascaded Multi-scale Feature Distillation (CMFD):} To preserve fine details, CMFD employs a multi-branch architecture with multi-scale convolutions and channel attention mechanisms to distll multi-scale details in order to capture text strokes and structures while reducing over-smoothing.
\end{itemize}

Finally, there is a reconstruction process to integrate BTFE and CMFD features via residual fusion for enhanced SCV output with stable gradients. Experimental results demonstrate that STM-Net achieves superior performance compared to state-of-the-art methods, particularly in scenarios involving rich text and scene changes.

\begin{figure*}[!t]
    \centering
    \includegraphics[width=\textwidth]{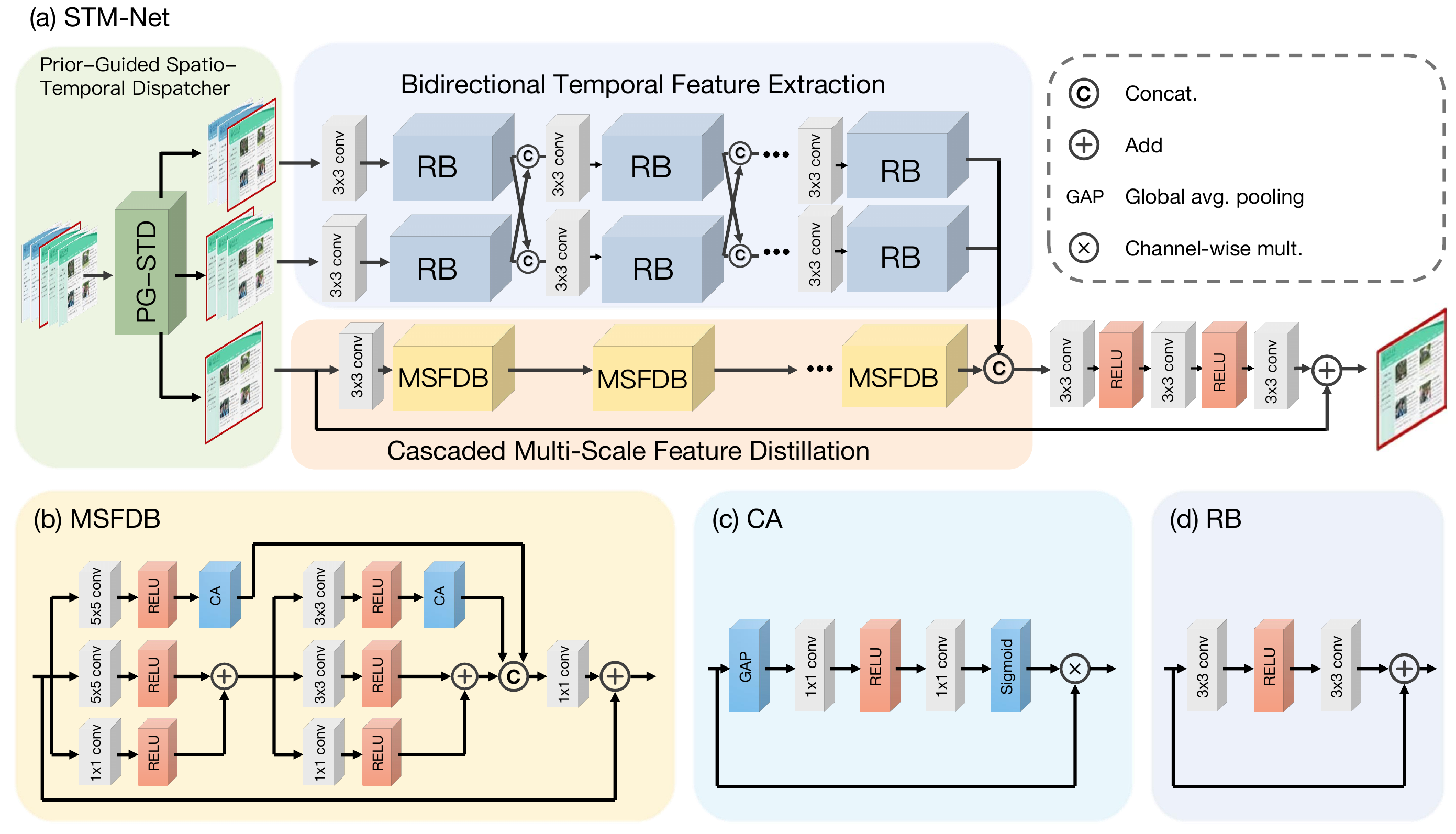}
\caption{The proposed STM-Net structure, comprising Prior-Guided Spatio-Temporal Dispatcher (PG-STD), Bidirectional Temporal Feature Extraction (BTFE) module, and Cascaded Multi-scale Feature Distillation (CMFD) module.}
\label{fig:stm_net_structure}
\end{figure*}

\section{Our Proposed STM-Net}

The proposed STM-Net framework, as illustrated in Fig. \ref{fig:stm_net_structure}, is designed to remove artifacts and address the challenges of SCV enhancement by utilizing a dual-stream architecture that efficiently handles abrupt motion and scene switch scenarios. We define $I_t^{LQ} \in \mathbb{R}^{H \times W}$ to represent a low-quality frame of size $H \times W$. The primary objective of STM-Net is to effectively enhance $I_t^{LQ}$ as the high-quality frame $\tilde{I}_t^{HQ} \in \mathbb{R}^{H \times W}$ by leveraging both temporal and spatial information from the current frame and a neighborhood of 2$R$ frames, which can be expressed as:
\begin{equation}
\tilde{I}_t^{HQ} = H_{STM-Net}(\{I_{t-R}^{LQ}, ..., I_t^{LQ}, ..., I_{t+R}^{LQ}\})
\label{eq:stmnet}
\end{equation}

As illustrated in Fig. \ref{fig:stm_net_structure}, the overall framework can be summarized as comprising three main modules: PG-STD, BTFE, and CMFD.

\subsection{Prior-Guided Spatio-Temporal Dispatcher (PG-STD)}

Unlike prior single-stream methods that risk feature contamination across scene switches, the proposed PG‑STD employs three parallel streams that process complementary temporal and spatial contexts to handle rapid changes in screen content. One stream processes the current frame with preceding frames $\{I_{t-2}^{LQ}, I_{t-1}^{LQ}, I_t^{LQ}\}$, another processes succeeding frames $\{I_t^{LQ}, I_{t+1}^{LQ}, I_{t+2}^{LQ}\}$, and a third stream operates on the current frame alone, $I_t^{LQ}$. The first two streams are fed into BTFE as dual parallel cross-connected streams that process different temporal input contexts to address rapid changes in screen content. These streams interconnect through a series of Residual Blocks (RBs) that extract hierarchical features by leveraging temporal dependencies through cross-connections. The single‑frame stream is fed into CMFD module to preserve fine spatial details and mitigate over‑smoothing.
% ==================================================================

\subsection{Bidirectional Temporal Feature Extraction (BTFE)}
BTFE addresses frequent scene switches in SCVs, where conventional single-stream networks average irrelevant future frames into $I_t$, causing artifacts.

It employs a symmetric dual-stream architecture, particularly with $R$=2 as in~\eqref{eq:stmnet}. One stream processes the preceding group $\{I_{t-2}^{LQ}, I_{t-1}^{LQ}, I_t^{LQ}\}$ to generate $F_{pre}$, while the other processes the subsequent group $\{I_t^{LQ}, I_{t+1}^{LQ}, I_{t+2}^{LQ}\}$ to generate $F_{post}$. Cross-connections via Residual Blocks (RBs) enable implicit prioritization:
\begin{equation}
\begin{aligned}
F_{pre}^n &= H_{RB}^n(F_{pre}^{n-1} + F_{post}^{n-1}) \\
F_{post}^n &= H_{RB}^n(F_{post}^{n-1} + F_{pre}^{n-1})
\end{aligned}
\end{equation}

Cross-stream additions enable the network to implicitly compare and prioritize the stable temporal context over corrupted streams during abrupt changes, avoiding explicit scene-cut detection overhead. This novel separation-and-fusion dynamically de-emphasizes irrelevant frames (e.g., post-switch $F_{pre}^n$ or pre-switch $F_{post}^n$), preserving high-frequency details in transitions like window dragging and switching, which is a key advance over traditional collective processing.

\subsection{Cascaded Multi-Scale Feature Distillation (CMFD)}
Deep SCV enhancement networks suffer feature degradation in deeper layers, which struggles to preserve high-frequency text edges and sharp boundaries, which cannot be addressed by natural-video multi-scale methods.

As shown in Fig. 1, each MSFD Block (MSFDB) uses multi-path architecture comprising parallel $1\times1$, $3\times3$, and $5\times5$ convolutional branches ($\Phi_{k\times k}$: conv + ReLU). Initial extraction on $F_{in}$ yields $f_1 = \Phi_{1\times1}(F_{in})$, $f_{5,1} = \Phi_{5\times5}(F_{in})$, with CA on the $5\times5$ branch giving $f_{5,2} = \text{CA}(f_{5,1})$; these fuse element-wise as $f_{fused}^1 = f_1 + f_{5,1}$.

To further distill spatial information, $f_{fused}^1$ undergoes secondary parallel processing: $g_1 = \Phi_{1\times1}(f_{fused}^1)$, $g_{3,1} = \Phi_{3\times3}(f_{fused}^1)$ (where $3\times3$ kernels capture mid-level context), refined by a second CA module to yield $g_{3,2}$. The final distilled output is generated by concatenating the multi-scale paths followed by a $1\times1$ projection:
\begin{equation}
    F_{distilled} = \Phi_{1\times1}([f_{5,2}, g_{3,2}, (g_1 + g_{3,1})])
\end{equation}
A global residual connection $F_{out} = F_{distilled} + F_{in}$ is applied to each MSFDB, facilitating stable gradient flow and feature reuse. This novel cascaded design iteratively distills hierarchical features via multi-scale kernels, uniquely balancing sharp text edges with uniform regions—mitigating deep-network degradation unlike conventional single-scale methods.

\subsection{Reconstruction}
Finally, BTFE temporal features and CMFD spatial details are concatenated, refined via three $3\times3$ convolutions, and added residually to $I_t^{LQ}$ to yield $\tilde{I}_t^{HQ}$. This novel fusion uniquely leverages their complementary strengths for detail-preserving SCV enhancement.

\begin{table*}[t]
\caption{$\Delta$PSNR ($\Delta$P) (dB) and $\Delta$SSIM ($\Delta$S) ($\times 10^{-3}$) at QP=37, average results for QP=32, 27, 22, and BD-Rate comparison.}
\label{tab:comparison}
\centering
\scriptsize
\setlength{\tabcolsep}{4pt}
\begin{tabular}{c|cc|cc|cc|cc|cc|cc|cc|cc|ccllc}

\cline{1-19}
\multirow{2}{*}{Seq. (QP=37)} & \multicolumn{2}{c|}{STDF-R3 \cite{deng2020spatio}} & \multicolumn{2}{c|}{QECF \cite{huang2022qecf}} & \multicolumn{2}{c|}{CAT \cite{liu2022content}} & \multicolumn{2}{c|}{CF-STIF \cite{luo2022coarse}} & \multicolumn{2}{c|}{STDR \cite{luo2023spatio}} & \multicolumn{2}{c|}{TGAF \cite{zhu2024compressed}} & \multicolumn{2}{c|}{EAST-LITE \cite{huang2024spatio}} & \multicolumn{2}{c|}{\textbf{STM-Net}} & \multicolumn{2}{c}{\textbf{STM-Net-L}} \\
 & \(\Delta\)P & \(\Delta\)S & \(\Delta\)P & \(\Delta\)S & \(\Delta\)P & \(\Delta\)S & \(\Delta\)P & \(\Delta\)S & \(\Delta\)P & \(\Delta\)S & \(\Delta\)P & \(\Delta\)S & \(\Delta\)P & \(\Delta\)S & \(\Delta\)P & \(\Delta\)S & \(\Delta\)P & \(\Delta\)S &  &  &  \\ \cline{1-19}
BigBuck & 0.327 & 3.18 & 0.325 & 3.23 & 0.318 & 4.19 & 0.369 & 3.98 & 0.408 & 4.33 & 0.432 & 4.11 & 0.411 & 4.82 & 0.438 & 4.61 & \textbf{0.485} & \textbf{5.04} \\
ChineseEditing & 0.273 & 2.22 & 0.244 & 1.63 & 0.200 & 1.24 & 0.360 & 2.41 & 0.321 & 1.77 & 0.48 & 3.79 & 0.525 & 4.17 & 0.544 & 3.68 & \textbf{0.65} & \textbf{4.68} \\
EnglishDoc & 0.867 & 2.78 & 0.770 & 2.70 & 0.951 & 3.27 & 1.243 & 3.41 & 1.266 & 4.01 & 1.101 & 3.63 & 0.991 & 3.36 & 1.194 & 3.46 & \textbf{1.275} & \textbf{4.04} \\
MissionControl1 & 0.492 & 4.17 & 0.503 & 4.12 & 0.477 & 4.00 & 0.625 & 4.72 & 0.546 & 4.59 & 0.721 & 5.73 & 0.647 & 6.02 & 0.760 & 5.94 & \textbf{0.782} & \textbf{6.39} \\
MissionControl2 & 0.569 & 5.33 & 0.563 & 5.19 & 0.568 & 5.18 & 0.672 & 5.67 & 0.689 & \textbf{6.08} & 0.677 & 5.38 & 0.612 & 5.55 & 0.700 & 5.48 & \textbf{0.748} & 5.94 \\
MissionControl3 & 0.545 & 5.06 & 0.551 & 4.96 & 0.535 & 4.98 & 0.625 & 5.25 & 0.591 & \textbf{5.51} & 0.61 & 4.3 & 0.573 & 4.56 & 0.666 & 4.59 & \textbf{0.724} & 5.08  \\
Paperpdf & 1.281 & 2.87 & 1.225 & 2.67 & 1.421 & 3.08 & 1.718 & 3.28 & 1.718 & 3.31 & 1.728 & 3.18 & 1.5 & 3.07 & 1.778 & 3.34 & \textbf{1.922} & \textbf{3.5} \\
Sephora & 0.779 & 2.38 & 0.831 & 2.34 & 0.864 & 2.79 & 1.127 & 3.32 & 1.189 & 3.81 & 1.233 & \textbf{4.24} & 1.069 & 3.85 & 1.165 & 3.51 & \textbf{1.273} & 3.88 \\
mixvideo & 0.301 & 3.46 & 0.365 & 3.51 & 0.329 & 3.05 & 0.278 & 3.85 & 0.306 & 4.16 & 0.516 & 4.13 & 0.528 & 4.35 & 0.558 & 4.45 & \textbf{0.577} & \textbf{4.9} \\
scSlideShow & 0.914 & 4.02 & 0.910 & 3.98 & 0.878 & 4.21 & 1.054 & 4.30 & 1.094 & 4.52 & 0.866 & 4.11 & 1.076 & 4.59 & 1.154 & 4.44 & \textbf{1.205} & \textbf{4.64} \\
scmap & 0.453 & 5.71 & 0.373 & 3.53 & 0.416 & 6.26 & 0.526 & 5.97 & 0.408 & 5.04 & 0.463 & 6.44 & 0.476 & \textbf{6.91} & 0.487 & 4.64 & \textbf{0.571} & 6.12 \\
scprogramming & 0.406 & 4.90 & 0.427 & 4.93 & 0.403 & 4.86 & 0.520 & 5.51 & 0.514 & 5.11 & 0.545 & 4.97 & 0.545 & 5.82 & 0.596 & 5.42 & \textbf{0.649} & \textbf{6.11} \\
scwebbrowsing & 1.008 & 3.28 & 0.907 & 3.38 & 0.969 & 3.56 & 1.286 & 3.93 & 1.046 & 3.72 & 1.137 & 3.78 & 1.107 & 3.69 & 1.329 & 3.93 & \textbf{1.48} & \textbf{4.09} \\ 
\cline{1-19}
\textbf{Avg. (QP=37)} & 0.632 & 3.80 & 0.615 & 3.55 & 0.641 & 3.90 & 0.800 & 4.28 & 0.777 & 4.30 & 0.808 & 4.45 & 0.774 & 4.67 & 0.875 & 4.42 & \textbf{0.949} & \textbf{4.95} \\ 
\cline{1-19}
\textbf{Avg. (QP=32)} & 0.533 & 2.09 & 0.531 & 2.12 & 0.541 & 2.04 & 0.704 & 2.54 & 0.655 & 2.37 & 0.656 & 2.19 & 0.684 & 2.37 & 0.752 & 2.37 & \textbf{0.801} & \textbf{2.58} \\
\textbf{Avg. (QP=27)} & 0.467 & 0.91 & 0.495 & 1.07 & 0.429 & 0.91 & 0.608 & \textbf{1.22} & 0.588 & 1.10 & 0.586 & 0.99 & 0.548 & 1.12 & 0.661 & 1.14 & \textbf{0.681} & 1.15 \\
\textbf{Avg. (QP=22)} & 0.417 & 0.53 & 0.470 & 0.54 & 0.426 & 0.55 & 0.533 & \textbf{0.64} & 0.537 & 0.62 & 0.550 & 0.61 & 0.496 & 0.61 & \textbf{0.563} & 0.60 & \textbf{0.563} & 0.62 \\
\cline{1-19}
\textbf{Avg. BD-Rate} & \multicolumn{2}{c|}{-5.24\%} & \multicolumn{2}{c|}{-5.38\%} & \multicolumn{2}{c|}{-5.20\%} & \multicolumn{2}{c|}{-6.48\%} & \multicolumn{2}{c|}{-6.23\%} & \multicolumn{2}{c|}{-6.46\%} & \multicolumn{2}{c|}{-6.43\%} & \multicolumn{2}{c|}{-7.12\%} & \multicolumn{2}{c}{\textbf{-7.46\%}} \\ 
\cline{1-19}
\end{tabular}
\end{table*}

\begin{figure}[t]
    \centering
    \includegraphics[width=\columnwidth]{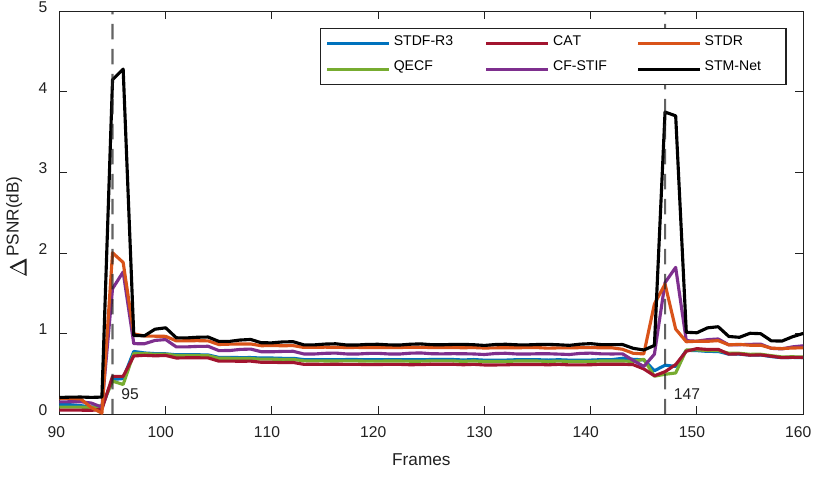}
    \caption{$\Delta$PSNR curves for \textit{mixvideo}; dashed lines mark scene switches.}
    \label{psnr_curve}
\end{figure}

\begin{figure}[t]
\centering
\includegraphics[width=\columnwidth]{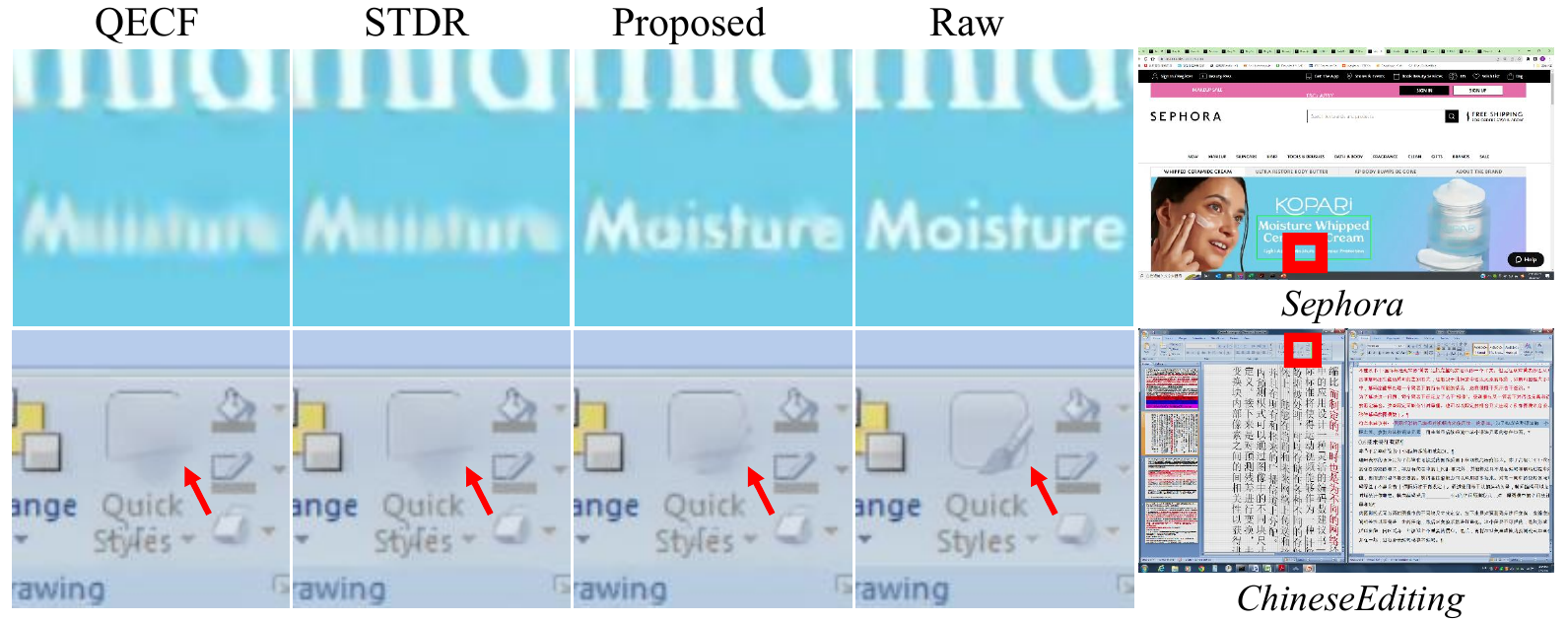}
\caption{Subjective visual quality comparison at QP=37. Top: Textual content enhancement on \textit{Sephora}. Bottom:  Graphical content enhancement on \textit{ChineseEditing}. STM-Net restores clearer texts and sharper edges compared to state-of-the-art methods.}
\label{fig:visual_comparison}
\end{figure}

\section{Experimental Results}
\subsection{Implementation Details}
We constructed a dataset of 41 SCV sequences (28 for training, 13 for testing), including standard Common Test Condition (CTC) sequences \cite{CTC}, SCVs from other sources, and self-captured SCVs containing abrupt motions \cite{huang2024spatio, tsang2019mode, jctvc2015screen}. Videos were encoded using the reference software HM16.20-SCM8.8 under the Low Delay Main SCC (LDMS) configuration at QPs 22, 27, 32, and 37.
The model was implemented in PyTorch and trained using the Charbonnier loss with the Adam optimizer (lr=$10^{-4}$) \cite{Adam} for 300,000 iterations.

\subsection{Objective Visual Quality Analysis}
We compare STM-Net against state-of-the-art methods, including STDF-R3 \cite{deng2020spatio}, QECF \cite{huang2022qecf}, CF-STIF \cite{luo2022coarse}, and STDR \cite{luo2023spatio}. As shown in Table \ref{tab:comparison}, STM-Net consistently achieves superior performance across all QP levels. At the highest compression (QP=37), STM-Net yields a $\Delta$PSNR of 0.875 dB, surpassing the second-best methods (TGAF and CF-STIF) by approximately 8.3--9.4\% and significantly outperforming STDF-R3 by 38.5\%. These gains are sustained at lower QPs. Furthermore, STM-Net achieves a BD-rate reduction of $-7.12\%$, representing a substantial improvement over CF-STIF ($-6.48\%$). Scaling up to STM-Net-L with more RB and MSFDB blocks further improves the BD-rate reduction to $-7.46\%$. Visual comparisons in Fig. \ref{fig:visual_comparison} confirm that STM-Net restores sharper edges and clearer text, validating its specialized design for screen content.

To evaluate the capability of our proposed STM-Net in handling scene switches, a screen content videos was selected to compute the $\Delta$PSNR curves for STDF-R3, QECF, CAT, CF-STIF, STDR, and our proposed method. The \textit{mixvideo} sequence is composed of spliced videos from CTC \cite{CTC}, which allows us to evaluate the performance of our method in scenarios involving abrupt scene transitions. The results are shown in Fig. \ref{psnr_curve}, where dashed lines indicate scene switch frames. The result demonstrates that our proposed method demonstrates an improvement during most of the transition points, highlighting its effectiveness in handling abrupt scene transitions. This robustness to screen content videos highlights the versatility and reliability of our method.

\begin{table}[!t]
\centering
\caption{Ablation Study of Different Architecture Components in STM-Net at QP=37}
\label{tab:ablation}
\resizebox{\columnwidth}{!}{\begin{tabular}{l@{\hspace{1em}}c@{\hspace{1em}}c@{\hspace{1em}}c@{\hspace{1em}}c}
\toprule
\textbf{Model Structure} & \textbf{STM-Net} & \textbf{STM-Nscale} & \textbf{STM-NHF} & \textbf{STM-Net-S} \\
\midrule
Multi-scale Feature Extraction & \checkmark & $\times$ & \checkmark & \checkmark \\
High-Frequency Processing & \checkmark & \checkmark & $\times$ & \checkmark \\
Number of RBs & 3 & 3 & 3 & 2 \\
Number of MSFDBs & 3 & 3 & 3 & 2 \\
\midrule
$\Delta$PSNR (dB) & 0.875 & 0.839 & 0.834 & 0.798 \\
\bottomrule
\end{tabular}}
\end{table}

\subsection{Subjective Visual Quality Analysis}
Fig. \ref{fig:visual_comparison} presents a visual comparison at QP=37. The top and bottom rows display the \textit{Sephora} and \textit{ChineseEditing} sequences, which contain small-sized text and computer-graphic icon, respectively. This is a challenging case where high-frequency information is heavily quantized. While competing methods produce blurry artifacts in regions containing text, STM-Net effectively reconstructs the character shapes. This capability is directly attributed to the CMFD module, where the 1$\times$1 convolution branch preserves fine details that are often lost by larger receptive fields.

\subsection{Ablation Study}
To verify the contribution of each architectural component, we conducted ablation experiments at QP=37, with the results summarized in Table \ref{tab:ablation}. Specifically, we compare the full model \textbf{STM-Net} (baseline with the complete architecture) against three ablated variants: \textbf{STM-Nscale}, which removes the multi-scale feature extraction paths (5$\times$5+CA and 3$\times$3+CA branches); \textbf{STM-NHF}, which removes the high-frequency processing paths (1$\times$1 + ReLU paths); and \textbf{STM-Net-S}, which reduces the network depth by using only 2 RB pairs and 2 MSFDBs.

\begin{figure}[tb] \centering
    \includegraphics[width=0.48\textwidth,height=0.16\textwidth]{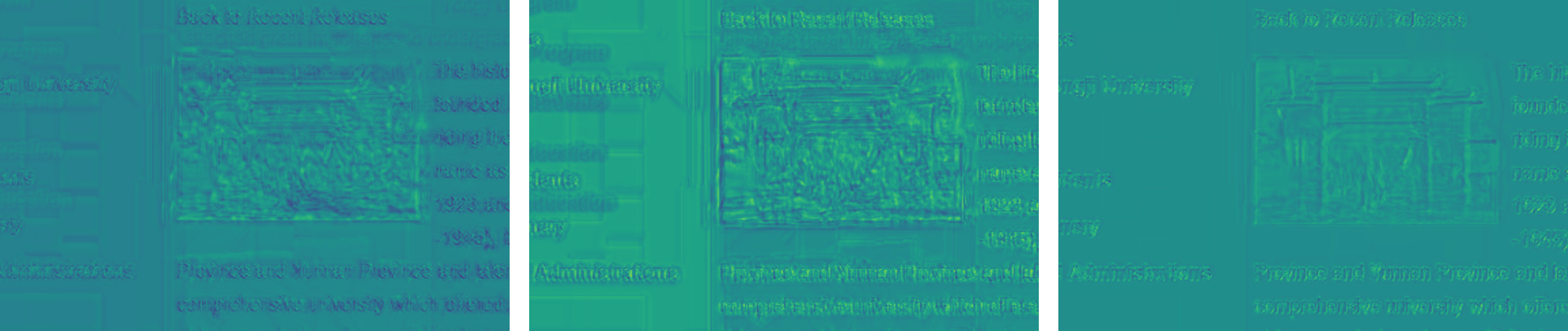}
    \caption{Visualization of Intermediate Features on Scene Switch. (\textbf{\textit{Left}}: BTFE Features Suppressed for Previous Frames; \textbf{\textit{Middle}}: BTFE Features Emphasized for Future Frames, e.g.: left menu changed; \textbf{\textit{Right}}: CMFD Features Emphasized for Screen Content Edges at Current Frame)} \label{fig:img1}
    \vspace{-1em}
    \label{fig:feature_visualization}
\end{figure}

\textbf{Impact of Multi-scale and High-frequency Processing:} Eliminating the multi-scale branches (STM-Nscale) or discarding the high-frequency 1$\times$1+ReLU paths (STM-NHF) both cause a consistent performance decrease of about 0.04~dB. This confirms that the multi-scale context modeling (via 5$\times$5+CA and 3$\times$3+CA) and the fine-grained high-frequency enhancement (via 1$\times$1+ReLU) are complementary and jointly important for handling SCVs.

\textbf{Impact of Network Depth:} Among all variants, reducing the overall depth (STM-Net-S: 2 RB pairs and 2 MSFDBs) yields the largest degradation (-0.077~dB). This suggests that sufficient depth is crucial for both BTFE and CMFD modules to learn complex temporal and spatial dependencies respectively, which are important for robust processing under scene switches and fine-grained high-frequency details.

\begin{table}[!t]
\centering
\caption{COMPARISON OF MODEL SIZES}
\label{tab:model_comparison}

\resizebox{\columnwidth}{!}{%
    \begin{tabular}{c|cccccccc}
    \hline
    Model & STDF-R3 & QECF & CAT & CF-STIF & STDR & \textbf{STM-Net-S} & \textbf{STM-Net} & \textbf{STM-Net-L} \\
    \hline
    $\Delta$ PSNR (dB) & 0.632 & 0.615 & 0.641 & 0.800 & 0.777 & 0.798 & 0.875 & 0.949 \\
    Parameters (KB) & 364.51 & 773.31 & 848.55 & 1242.10 & 1521.13 & 674.37 & 1009.75 & 1345.13 \\
    \hline
    \end{tabular}%
}
%\vspace{-2.8em}
\end{table}

\subsection{Model Scaling}
Table \ref{tab:model_comparison} compares the average $\Delta$PSNR against the model parameters. These results are averaged over all test sequences. As a result, the performance of STM-Net significantly surpasses other methods and requires fewer model parameters than CF-STIF and STDR, as in Table \ref{tab:model_comparison}. In addition, our STM-Net is a modular network, allowing for easy model scaling by varying the number of RB and MSFDB blocks. Therefore, in applications with computational limitations, we can use a lightweight structure, such as STM-Net-S, with fewer blocks ($N=M=2$). STM-Net-S requires fewer model parameters than STDF-R3, QECF, CAT, and STDR, as shown in Table \ref{tab:model_comparison}, yet still achieves higher $\Delta$PSNR of 0.798 dB. This highlights the efficiency and effectiveness of our proposed method. On the other hand, scaling up as STM-Net-L ($N=M=4$) yields further improvements of nearly 1 dB (0.949 dB), which demonstrates that model scaling is applicable for our STM-Net on SCV enhancement.

\subsection{Feature Visualizations}
Fig. \ref{fig:feature_visualization} presents intermediate feature visualizations for BTFE and CMFD using Grad-CAM \cite{selvaraju2017gradcam}, which validates their key advantages. These features illustrate how BTFE's dual-stream design automatically prioritizes the more temporally consistent direction during abrupt scene switches, while CMFD effectively preserves sharp text strokes and high-frequency details. In BTFE visualizations, light colors indicate high activation weights assigned to the more temporally consistent stream (reliable frames), while dark colors denote suppressed weights on disrupted streams during abrupt scene switches, which demonstrates automatic prioritization without explicit detection. Meanwhile, CMFD features (right) preserve sharp text strokes and high-frequency details through multi-scale distillation, with light-to-dark gradients highlighting enhanced edge recovery and reduced over-smoothing compared to inputs.

\section{Conclusion}
This paper presents STM-Net, a specialized framework for screen content video quality enhancement. By seamlessly integrating Prior-Guided Spatio-Temporal Dispatcher (PG-STD), bidirectional temporal feature extraction (BTFE) with cascaded multi-scale feature distillation (CMFD), it effectively tackles SCV-specific challenges such as abrupt scene switches, discontinuous motions, and high-frequency detail loss from compression artifacts. Extensive experiments confirm that STM-Net achieves state-of-the-art performance, offering a promising solution for high-quality screen content delivery.

\bibliographystyle{IEEEtran}
\bibliography{refs}
\end{document}